\documentclass[10.7pt,]{article}

\usepackage[letterpaper, margin=2.54cm, top=2.54cm]{geometry}
\usepackage[super,comma,sort&compress]{natbib}
\usepackage{lmodern}
\usepackage{authblk} 
\usepackage{amssymb,amsmath}
\usepackage{wrapfig}
\usepackage{graphicx,grffile}
\usepackage[labelfont=bf,labelsep=period]{caption}
\usepackage{ifxetex,ifluatex}
\usepackage{algpseudocode}
\usepackage{fixltx2e} 
\ifnum 0\ifxetex 1\fi\ifluatex 1\fi=0 
  \usepackage[T1]{fontenc}
  \usepackage[utf8]{inputenc}
\else 
  \ifxetex
    \usepackage{mathspec}
  \else
    \usepackage{fontspec}
  \fi
  \defaultfontfeatures{Ligatures=TeX,Scale=MatchLowercase}
\fi
\IfFileExists{upquote.sty}{\usepackage{upquote}}{}
\IfFileExists{microtype.sty}{%
	\usepackage{microtype}
	\UseMicrotypeSet[protrusion]{basicmath} 
}{}

\usepackage{placeins}
\usepackage{float}
\usepackage{algorithm}

\usepackage[compact]{titlesec}
    \titlespacing{\section}{0pt}{2ex}{1ex}
    \titlespacing{\subsection}{0pt}{1ex}{0ex}
    \titlespacing{\subsubsection}{0pt}{0.5ex}{0ex}

\usepackage{sectsty}
\allsectionsfont{\fontsize{10}{10}\selectfont}
\subsectionfont{\itshape\bfseries\fontsize{10}{10}\selectfont}
\subsubsectionfont{\normalfont\itshape}

\makeatletter
\renewcommand\subsubsection{\@startsection{subsubsection}{3}{\z@}%
	{-3.25ex\@plus -1ex \@minus -.2ex}%
    {-1.5ex \@plus -.2ex}
    {\normalfont\itshape}}
\makeatother

\makeatletter 
\renewcommand\@biblabel[1]{#1.} 
\makeatother %

\usepackage{etoolbox}
\makeatletter
\patchcmd{\@maketitle}{\LARGE}{\bfseries\fontsize{15}{16}\selectfont}{}{}
\makeatother

\makeatletter
\def\maxwidth{\ifdim\Gin@nat@width>\linewidth\linewidth\else\Gin@nat@width\fi}
\def\maxheight{\ifdim\Gin@nat@height>\textheight\textheight\else\Gin@nat@height\fi}
\makeatother

\setkeys{Gin}{width=\maxwidth,height=\maxheight,keepaspectratio}
\ifx\paragraph\undefined\else
\let\oldparagraph\paragraph
\renewcommand{\paragraph}[1]{\oldparagraph{#1}\mbox{}}
\fi
\ifx\subparagraph\undefined\else
\let\oldsubparagraph\subparagraph
\renewcommand{\subparagraph}[1]{\oldsubparagraph{#1}\mbox{}}
\fi

\usepackage{tcolorbox}

\definecolor{bg}{gray}{0.95}

\usepackage{hyperref}
\hypersetup{
	unicode=true,
	pdftitle={My Cool Title Here},
	pdfauthor={Author One, Author Two, Author Three},
	pdfkeywords={keyword1, keyword2},
	pdfborder={0 0 0},
	breaklinks=true
}
\usepackage{cleveref}

\title{\vspace{-2em} MentalHealthAI: Utilizing Personal Health Device Data to Optimize Psychiatry Treatment}
\author[ ]{\bf\fontsize{13}{14}\selectfont Manan Shukla and Oshani Seneviratne, PhD\vspace{-.7em}}
\affil[ ]{\bf\fontsize{13}{14}\selectfont Rensselaer Polytechnic Institute, Troy, NY, USA}
\date{} 

\begin{document}
\maketitle
\vspace{-4em} 

\section{Abstract}\label{abstract}
Mental health disorders remain a significant challenge in modern healthcare, with diagnosis and treatment often relying on subjective patient descriptions and past medical history. To address this issue, we propose a personalized mental health tracking and mood prediction system that utilizes patient physiological data collected through personal health devices. Our system leverages a decentralized learning mechanism that combines transfer and federated machine learning concepts using smart contracts, allowing data to remain on users' devices and enabling effective tracking of mental health conditions for psychiatric treatment and management in a privacy-aware and accountable manner. We evaluate our model using a popular mental health dataset that demonstrates promising results. By utilizing connected health systems and machine learning models, our approach offers a novel solution to the challenge of providing psychiatrists with further insight into their patients' mental health outside of traditional office visits.
\section{Introduction}\label{introduction}
Mental health conditions such as depression and anxiety are some of the most challenging medical problems to diagnose and treat. Current treatment guidelines for these disorders primarily utilize subjective assessments, relying on patient self-report or clinician evaluation to inform clinical decisions. As such, the lack of objective markers for clinical outcomes presents a significant bottleneck in psychiatry. Furthermore, a patient's mood or emotions may change over time, but clinicians only have access to a patient’s data at the time of the visit, leading to a potentially biased sampling of the patient’s mental state. To address this, collecting data from the patient over a long period would be ideal for effective diagnosis and treatment. However, collecting such data is also challenging due to privacy concerns.
Connected health applications enable data to be generated and stored in a decentralized manner, where the data may reside cross-device.
A common challenge in health informatics in federated and decentralized settings is that training and test data are not independently and identically distributed (non-IID), which is especially true in scenarios that apply to predict the mental health of individuals using a combination of medical and environmental signals.
Because health data is typically not identically distributed, the generalization performance tends to be worse, and lower accuracy can result from overlooking the distribution shift in the training and testing data\cite{zhao2018federated}.
More importantly, since non-IID data in healthcare applications comes from different clients, protecting data privacy is crucial in decentralized learning settings\cite{rieke2020future}.

Furthermore, applying connected health technologies in a mental health population poses multiple problems\cite{ivanova2020mental}. First is the concern about data security and privacy. 
Studies have shown that mental health populations typically consider their data sensitive and vary in sharing this information due to perceived mental health stigmas. Surveys have shown that 65\% of patients with mental health disorders are unlikely to share patient data with their psychiatrists\cite{grando2020mental}. 
If the psychiatrists aim to rely on patients' history, studies\cite{vermani2011rates} have shown patient histories only to be 62\% accurate, leading to psychiatric misdiagnoses as high as 65.9\% for major depressive disorders and 85.8\% for panic disorders.
Therefore, a technological solution is necessary to provide psychiatrists with health insights without collecting raw data from the patient's smart health devices. Second, current models do not account for the granularity of mental health disorders. As explained in the American Psychiatric Association's Clinical Practice Guidelines\cite{cpg}, patient emotions are subject to rapid changes within the span of a day or a week, and elements such as sleep or diet can lead to quick changes in mood. While many have utilized information from Electronic Health Records (EHR) to predict mental health crises\cite{garriga2022machine}, these models overlook granular patient changes. Therefore, they cannot generate a patient baseline (in fact, getting data through facial expressions or EHR systems can lead to biased results). Understanding the immediate effects of medication, such as antidepressants, is crucial for psychiatrists and requires granular patient data that cannot be retrieved otherwise. Currently, the most feasible way to collect this granular patient data is through a smartphone and a patient's health devices. This method, however, has the issue of unequal data streams. Different patients have different personal health devices. For example, while one patient may have five devices, another may only have one. While training a model on the patient with five devices may lead to better results, the input from a patient population to this model will decrease (as not many patients have so many personal health devices). Therefore, there is a need to obtain insights even with the feature types being unequal from patient to patient. 
We present a decentralized federated learning algorithm called \emph{MentalHealthAI} to alleviate these challenges. 
First, \emph{MentalHealthAI} uses on-device machine learning to prevent data from leaving the patient's smartphone. However, smart contracts are utilized in this framework to elect an aggregator, thereby creating a decentralized aggregator instead of a traditional centralized server.
A self-executing piece of code, called a smart contract, can encode rules that will be executed in a decentralized manner on a blockchain\cite{zheng2020overview}.
The data remains on the patient's device in each epoch, and the model parameters are transferred from that device to the aggregator. Second, as each smartphone may collect a different set of patient features, \emph{MentalHealthAI} utilizes a decision tree based methodology to derive model insights even when features and labels are not necessarily uniform. 

\section{System Design and Implementation}
\label{sec:design}

\begin{wrapfigure}{r}{0.58\textwidth}
  \vspace{-70pt}
  \begin{center}
    \includegraphics[width=0.56\textwidth]{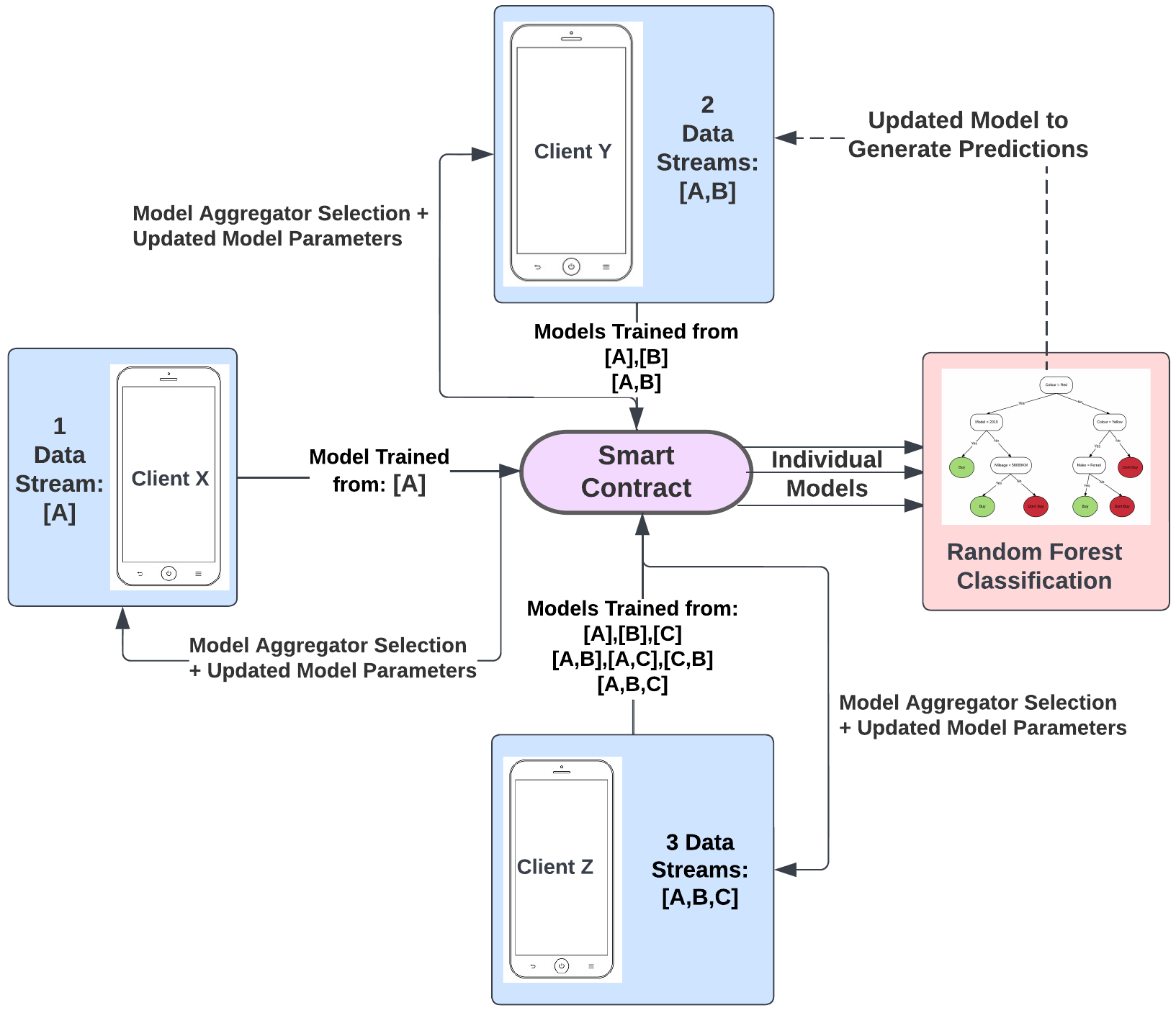}
  \end{center}
  \vspace{-20pt}
  \caption{System Architecture}
  \label{fig:architecture}
  \vspace{-10pt}
\end{wrapfigure}


At its core, the current system is a decentralized-learning infrastructure that utilizes physiological data to predict patient moods and therefore provides mental insights to a patient's psychiatrist without the requirements of a uniform set of features (as is necessary for typical machine learning algorithms). The overall architecture can be found in \Cref{fig:architecture}, and its specific features are described below. Clients X, Y, and Z each represent different patients. Each patient owns several IoT devices, indicated by the different data streams A, B, and C. Each data stream has the same dimensions but different content (A may be heart rate data, B may be blood pressure, and C may be skin temperature). The final model is created by adding the union of models trained from different combinations of data streams to the random forest decision tree classification system. During the evaluation, we use the features in the POPANE dataset\cite{behnke2022psychophysiology} as individual data streams to simulate this concept.


Personal health data do not contain uniform features from patient to patient. The common problem is that the devices used by different individuals are different. 
Traditional machine learning is limited in cases where one patient has data collected from many disparate data streams, such as heart rate, blood pressure, electroencephalogram (EEG), and electrocardiogram (ECG), while another patient has only one (for example, just the heart rate). This limitation exists because only the intersection of feature types is considered rather than every feature type present. 
For each patient, we assume a smartphone acts as the gateway between the patient's IoT devices, as depicted in \Cref{alg:iot-fed}.

\begin{algorithm}[!htbp]
\caption{Smartphone (Client) Federation of IoT Devices}\label{alg:iot-fed}
\begin{algorithmic}[1]
\State $DeviceAddresses = [DevAddr1,DevAddr2...]$ \Comment{IoT device addresses}
\State $MentalHealthAIdf = dataframe()$
\For{$addr \in \{1,\dots,DeviceAddresses\}$} \Comment{Collect data from each IoT device within private mobile environment}
    \State{$data \gets GetData(addr)$}
    \State $MentalHealthAIdf.AddColumn(data)$
\EndFor
\end{algorithmic}
\end{algorithm}

We utilized the  POPANE dataset\cite{behnke2022psychophysiology} for the simulation, where we divided the patient population based on the type and number of data streams each patient has. For example, we place patients with six of the same data streams (set $A$) in a different cohort than patients with only three data streams (set $B$). Here, \emph{data streams} can refer to heart rate and blood pressure data. However, if set B is a subset of set A, $A \cap B$ can be added to cohort B's training set (as the data streams used in B and $A \cap B$ are the same). 
Now, consider a patient population where the number of data streams a patient has varies from 1 to 6. For any patient with >1 data streams, a power set excluding the empty set is generated as shown in \Cref{alg:data-streams}. For example, given a patient with data streams d = \{A, B, C\}, P(d) = [\{\}, \{A\}, \{B\}, \{A, B\}, \{C\}, \{A, C\}, \{B, C\}, \{A, B, C\}], excluding \{\}, we divert each element in this power set (representing a set of data streams) into separate cohorts. This process maximizes the utility of patients with multiple data streams, as it maximizes the amount of data found in each cohort (in comparison to simply dividing the patient population based on the number of data streams present). We then train multiple machine learning models on these cohorts, where one model is trained from data from one cohort. Note that the labels are unaltered, regardless of the feature subset. When combined with \emph{MentalHealthAI}'s decentralized AI architecture, it is also important to note that multiple smartphones will be selected as aggregators but will be training different models with different training subsets. 

\begin{algorithm}[!htbp]
\caption{Smartphone (Client) Generation of Data Stream Combinations}\label{alg:data-streams}
\begin{algorithmic}[1]
\State $n = MentalHealthAIdf.headers.length()$
\State $subsets = [] $
\State $n = 0 $
\While{$n \neq 1$} \Comment{Generate all subsets of MentalHealthAIdf features with sizes $ \{n-1,\dots,1\}$}
    \State $SubsetList = GenerateSubsets(MentalHealthAIdf,SubsetSize=n)$
    \State $subsets.append(SubsetList)$
    \State $subsets.flatten() $
    \State $n = n-1$ \Comment{"subsets'' is now a one-dimensional list with every possible subset from size $ \{1,\dots,n\}$}
\EndWhile
\end{algorithmic}
\end{algorithm}

Based on the patient population and available data streams at a given time, certain models will be more accurate than others, and this relationship can change frequently. Furthermore, every patient's mood with different baseline emotion levels may differ. While models can predict a large portion of the population successfully, they may not be accurate enough for a specific patient. \Cref{alg:model-gen} shows the model generation process for each client based on the available datasets.

\begin{algorithm}[!htbp]
\caption{Smartphone (Client) Model Generation}\label{alg:model-gen}
\begin{algorithmic}[1]

\State $models = [] $ \Comment{For models trained on each subset in subsets}
\For{$subset \in \{subsets\}$}
    \State $model = GenerateModel(subset)$ \Comment{Generate a model with the subset as the features and the original labels as the label. See \Cref{fig:BNN}. Note: Model Generation/Training can happen in parallel.}
    \State $models.append(model)$
\EndFor
\end{algorithmic}
\end{algorithm}

Using a smart contract deployed on a blockchain as the ``secure model aggregator,'' the client interacts with it by emitting events to indicate learning has finished. The corresponding smart contract code is depicted in \Cref{alg:scpseudo}. The smart contract employs a voting process to elect the next ``leader'' to perform model aggregation. 
As each model has been trained on different feature subsets, decentralized aggregation occurs independently for each model. For example, if we have three models trained on features [A], [A,B], and [A,B,C], each of these models will be aggregated with other models trained on the same set of features from a different patient on a different client.
If the patient's smartphone is not elected as an aggregator, the smart contract will send the model parameters from the patient's smartphone to the smartphone elected as the model aggregator.

\begin{algorithm}[!htbp]
\caption{Smart Contract Leader/Aggregator Election}\label{alg:scpseudo}
\begin{algorithmic}[1]
\State $AggregatorStorage = \{\}$ 


\State $AggreagatorStorage = InitializeAggregatorStorage(models,DataStreams)$ 

\For {$(model,DataStream) \in \{AggregatorStorage\}$} 
    \State $VotingResultAddress = SelectAggregator(DataStream)$ 
    \State $EmitEvent(VotingResultAddress, DataStream)$


        
\EndFor

\State $TransactData(AggregatorStorage,SmartphoneAddress)$

\end{algorithmic}
\end{algorithm}

The clients will interact with the smart contract as shown in \Cref{alg:model-agg}.
Once the clients have finished training, they will notify the smart contract and be considered for the next ``leader'' election.
The client will also monitor events emitted from the smart contract to see if it is elected as an aggregator. If it is, then it will receive models from other smartphones.

\begin{algorithm}[!htbp]
\caption{Smartphone (Client) Interaction with the Smart Contract for Model Aggregation}\label{alg:model-agg}
\begin{algorithmic}[1]

\State $Transact(SetLearningFinished(True),SmartContractAddress)$ \Comment{Indicate learning finished.}
\State $Transact(GetModelParams(models,subsets),SmartContractAddress)$ \Comment{Transmit  Model Parameters.}
\State $AssignedDataStreams = []$ 
\If{$MonitorAggregatorEvent(SmartContractAddress,SmartphoneAddress)$} 
    \State $AssignedDataStreams = Transact(GetAssignedDatastream,SmartContractAddress)$
    \State $models = CollectNodeModels()$ 
\EndIf
\end{algorithmic}
\end{algorithm}

\begin{figure}[!htbp]
    \centering
    \includegraphics[width=\columnwidth]{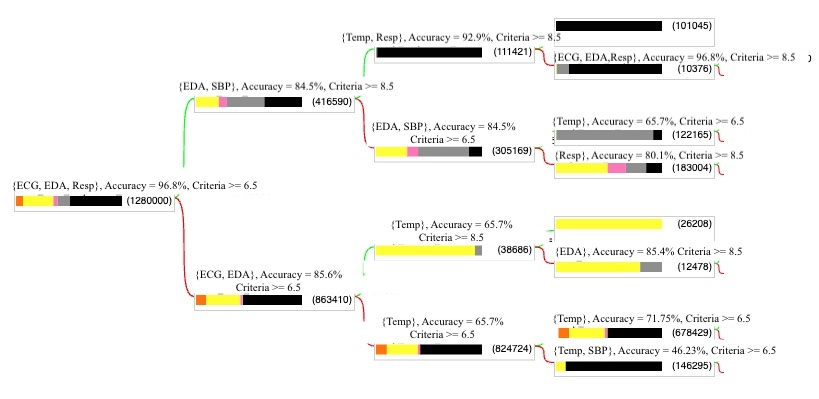}
    \caption{An Example Decision Tree. Each branch title is in the following format: data streams utilized in curly braces, the local model accuracy, and the criteria to split the branch. Below, the multi-colored branches represent the label distribution in the model, and the number in parentheses represents the total feature/labels utilized. Please see \Cref{fig:BNN} to cross-reference the acronyms used in this diagram.}
    \label{fig:DTree}
\end{figure}

Once the model parameters have been received, utilizing a decision tree, the ``leader'' client select the best prediction model for the patient as shown in \Cref{alg:model-agg-decision-tree}. 
It collects mapping from the smart contract with data stream as \emph{key} and aggregator smart contract address as \emph{value}.
For example, assume a patient has three devices/data streams. This patient's models include every model trained on the following data stream combinations: [\{A\}, \{B\}, \{A, B\}, \{C\}, \{A, C\}, \{B, C\}, \{A, B, C\}].
Then, a calibration period is set for a certain period to collect new patient emotional features/labels (in \Cref{alg:model-agg-decision-tree}, it is set to 7 days). Each set of features in our simulation contributes to the models generated daily. The random forest decision tree (such as the one shown in \Cref{fig:DTree}) will then use these models to predict the patient's emotional labels. The decision tree is run on the patient's smartphone after data stream based models have been generated and distributed back to the individual nodes from the aggregator. 

\begin{algorithm}[!htbp]
\caption{Smartphone (Client) Local Model Aggregation Using Decision Tree}\label{alg:model-agg-decision-tree}
\begin{algorithmic}[1]



\State $ModelStorage = \{\}$ 

\State $AggregatorStorage = GetAggregatorStorage(SmartContractAddress)$ 

\For{$subset \in \{subsets\}$} 
    \State $model = GetModel(subset,AggregatorStorage[subset])$ 
    \State $ModelStorage[subset] = model$
\EndFor
\State $CurrentData = dataframe()$ 
\For {$7 Days Time...$}:
    \State $CurrentData.AddRow(PhysiologicalData,EmotionalState)$
\EndFor
\State $DecisionTreeData = dataframe()$

\For {$Feature \in \{PhysiologicalData\}$} 
    \State $DecisionTreeData.AddCol(features)$
\EndFor
\State $DecisionTreeData.AddCol("EmotionalLabel")$
\For {$model \in \{ModelStorage\}$}
    \State $DecisionTreeData.AddCol(model.ModelName)$ 
    \For {$PhysiologicalData \in \{CurrentData\}$}
        \State $DecisionTreeData.AddRow(PhysiologicalData,EmotionalLabel,ModelPrediction)$ 
    \EndFor
\EndFor
\State $DecisionTreeModel = TrainDecisionTree(DecisionTreeData)$ 
\end{algorithmic}

\end{algorithm}

\section{Evaluation and Results}\label{methods}


\begin{wrapfigure}{r}{0.5\textwidth}
  \vspace{-70pt}
  \begin{center}
    \includegraphics[width=0.48\textwidth]{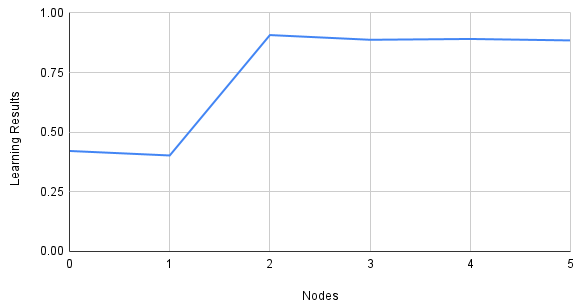}
  \end{center}
  \vspace{-20pt}
  \label{fig:learning_results_vs_nodes}
  \caption{Learning Results After Leader Election and Model Aggregation. \emph{Nodes} refer to the other smartphones contributing to the combined model.}
  \vspace{-10pt}
\end{wrapfigure}

\begin{wrapfigure}{r}{0.5\textwidth}
  \vspace{-20pt}
  \begin{center}
    \includegraphics[width=0.48\textwidth]{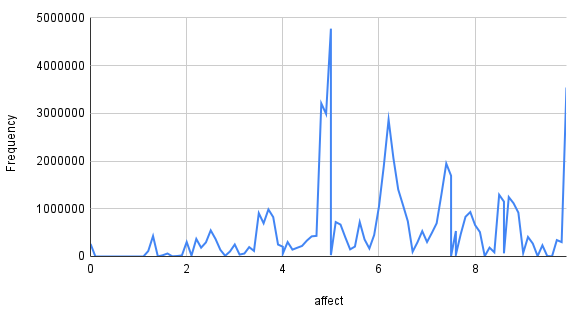}
  \end{center}
  \vspace{-20pt}
    \caption{Frequency of Various Affects in the POPANE Dataset\cite{behnke2022psychophysiology}}
    \label{fig:FreqvsAffect}
  \vspace{-10pt}
\end{wrapfigure}

We evaluated our system using a mental health dataset named POPANE\cite{behnke2022psychophysiology}. The POPANE dataset contains a set of 142 patients whose ECG, Electrodermal Activity (EDA), Skin Temperature (ST), Respiration (Resp), Systolic Blood Pressure (SBP), and Diastolic Blood Pressure (DBP) have been measured and labeled with positive and negative affect, which is rated from a scale of 0-10, with 0 indicating negative affect, and 10 indicating positive affect. We chose this dataset primarily because it closely matches our use case. A personal health device can measure each of the physiological parameters given above, and training on such a dataset can provide insight into the utility of such a system in a much larger population. Secondly, the data provided is collected on a second-to-second basis, similar to the collection rates found in many current IoT devices, such as smartwatches that measure heart rate or ECGs on the patient's skin. Finally, a major advantage of utilizing the POPANE dataset is its non-IID distributed data, as seen in \Cref{fig:FreqvsAffect}. The figure clearly shows that the affect is not equally distributed throughout the dataset (and is not likely to represent the standard population), which is more akin to what may be present in real-world situations, where random samples proportionate to the overall population are unlikely. Thus, through this dataset, we aim to investigate the resilience of \emph{MentalHealthAI} in non-IID settings. 


\begin{wrapfigure}{r}{0.8\textwidth}
  \vspace{-40pt}
  \begin{center}
    \includegraphics[width=0.78\textwidth]{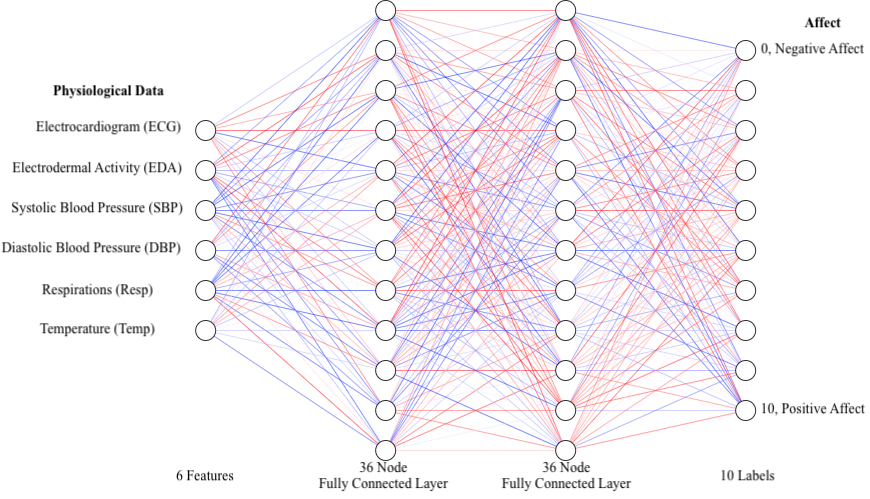}
  \end{center}
  \vspace{-20pt}
    \caption{A Simplified View of the Neural Network Model Architecture}
    \label{fig:BNN}
  \vspace{-10pt}
\end{wrapfigure}

First, we assessed the training results from a model run on a centralized server. The model was a simple Artificial Neural Network (ANN) with three dense layers with softmax activation, as shown in ~\Cref{fig:BNN}. We decided upon the activation function based on favorable learning results. As mentioned, we used six physiological features to assess the patient's affect, ranked from 0-10 to serve as the output. Each physiological feature is considered a separate data stream for this evaluation, containing data from different IoT devices. We set the data into a train-test split of 70-30\% and used a sparse categorical cross-entropy loss function due to the nature of the output categorical labels. Multiple checkpoints were implemented, such as early stopping (which will stop training if accuracy does not improve after multiple epochs in a row) and learning adjustment (which will lower the learning rate by a factor of ten if the accuracy does not improve). We ran the model for 107 epochs and stopped the learning process because there was no change in training accuracy. After multiple trials, this epoch value led to the best learning result in our model. The overall accuracy was approximately 86\%.


Based on these results, we can conclude that there is a link between physiological parameters and a patient's emotional state. We chose accuracy as our primary evaluation metric to ensure the model is clinically viable. 
We then evaluated the decentralized learning aspects of this system. Since we could not acquire physical devices to test the model's performance in the real world, we evaluated the decentralized learning components through simulation. In this simulation, we assume a consortium of 142 patients modeled using the POPANE dataset, each with data collected through IoT devices. A global model updates itself based on data from each patient to form the final trained model. We trained the models in the same fashion as the ANN discussed above.  \Cref{fig:learning_results_vs_nodes} shows the test-set accuracy after training on each node.
As shown in \Cref{fig:learning_results_vs_nodes}, successful learning can happen in a discontinuous situation. While the nodes had an initial training accuracy of 51\%, this increased immediately to 86\% after training from two additional nodes, confirming that \emph{MentalHealthAI} can obtain high accuracy even in distributed settings. However, note that such results may not be obtainable in real-world conditions. Primarily, data collected in the POPANE dataset has been obtained in a controlled environment rather than during regular day-to-day activities. Therefore, if truly deployed in a community, there is a greater chance of false positives, false negatives, and inaccurate readings from IoT devices. However, given accurate input data, we assert that \emph{MentalHealthAI} can be deployed in such a community setting.
We then evaluated the decision tree aspect of the system to determine the model accuracy in non-ideal settings. We divided the 142 patients in the POPANE dataset into four patient cohorts. Each cohort represented patients with a certain number of IoT devices (1 device, 2 devices, 3 devices, or 4 devices). Similar to what we explained in the methods section, we extracted data streams from each cohort based on the power set of their features. For example, in the cohort with 2 data streams (A and B), three sets of data were created: {A, B, (A, B)}, each with the same label. We repeated this process for each patient cohort. Models were trained on the following data streams: {ST}, {ECG}, {ST, ECG}, {EDA}, {ST, EDA}, {ECG, EDA}, {ST, ECG, EDA}, {Resp}, {ST, Resp}, {ECG, Resp}, {ST, ECG, Resp}, {EDA, Resp}, {ST, EDA, Resp}, {ECG, EDA, Resp}, {ST, ECG, EDA, Resp}. Next, we simulated the ``calibration'' period, where each model generated emotion predictions based on new data to which the models were not exposed. This data then served as the input to the random forest model, which then provided the emotional predictions based on the predictions of the previously trained models. \Cref{fig:DTree} depicts the decision tree for a single client (i.e., a smartphone belonging to a patient). 

We simulated a standard baseline solution, \emph{MentalHealthAI-Baseline}, to the above problem as a means of comparison. As a typical machine learning model cannot utilize different feature sets, the most optimized results will likely only come from the cohort with 4 data streams (35 patients). We trained standard ANN with the same hyperparameters as above on this data set, with an overall accuracy of 86\%. 
In comparison, \emph{MentalHealthAI-Fed} had an overall accuracy of 80\%, a substantial improvement. By utilizing unequal feature sets through multiple model combinations and a random forest model, one can improve learning results compared to a model that requires uniform features. While this accuracy level is lower than the original baseline model, it is important to acknowledge the differences in data. The baseline ANN model simulated an ideal world where 142 willing patients with access to 6 separate IoT devices. However, finding 142 patients with more than three personal health devices in the real world is intuitively infeasible for many reasons, such as cost and access. However, through this unique \emph{MentalHealthAI} framework, we demonstrate that high accuracy is achievable even in less-than-realistic settings. We believe that this occurs due to multiple reasons. First, \emph{MentalHealthAI} utilizes models without noise and irrelevant features, making them less susceptible to their effects. Second, models trained on less number of features can succeed by having access to a greater number of patients. Third, a random forest model can select the best model for the patient, a choice that can change over time. 
Finally, \emph{MentalHealthAI} was compared to current state-of-the-art emotion prediction systems and machine learning methods in adjacent domains. As shown in \Cref{tbl:CompareAlgorithms}, it is clear that compared to other past AI models, \emph{MentalHealthAI} can produce greater accuracy with both the baseline ANN model and the decentralized decision tree architecture. Note that due to the novelty of the POPANE dataset at the time we developed our model, we were unable to compare our results to similar models that may have been trained on the same data. 

\begin{table}[]\label{tbl:CompareAlgorithms}
\centering
\caption{Comparison of Mental Health Prediction Systems\\
\textit{\textbf{Legend: }
Diastolic Blood Pressure (DBP),
Electrocardiogram (ECG), 
Electrodermal Activity (EDA)
Electroencephalogram (EEG), 
Environmental Temperature (ET),
Fingertip Blood Oxygen Saturation (OXY),
Galvanic Skin Response (GSR),
Heart Rate (HR),
Heart Rate Variability (HRV),
Respirations (Resp),
Skin Conductivity (SC), 
Skin Temperature (ST), 
Systolic Blood Pressure (SBP),
}}
\begin{tabular}{|l|l|r|}
\hline
\textbf{Algorithm/Model}          & \textbf{Data Used}                                                                                                                               & \multicolumn{1}{l|}{\textbf{Accuracy}} \\ \hline
Lan Z. et al\cite{lan2016real}                      &  EEG                                                                                                                             & 49.63\%                                \\ \hline
Cheng et. al\cite{8304293}                      & \begin{tabular}[c]{@{}l@{}}ECG, HRV \end{tabular}                                                         & 79.51\%                                \\ \hline
Peter et. al\cite{peter2005wearable}                      & \begin{tabular}[c]{@{}l@{}}ST, SC, ET, HR\end{tabular}      & 75\%                                   \\ \hline
Wen et. al\cite{Wen2014EmotionRB}                        & \begin{tabular}[c]{@{}l@{}}GSR, OXY, HR\end{tabular}                 & 74\%                                   \\ \hline
\textbf{MentalHealthAI-Baseline}                        & \begin{tabular}[c]{@{}l@{}}\textbf{ECG, EDA, ST, Resp, SBP, DBP}\end{tabular} & \textbf{86\%}                                   \\ \hline
\textbf{MentalHealthAI-Fed}                   & \begin{tabular}[c]{@{}l@{}}\textbf{ECG, EDA, ST, Resp, SBP, DBP}\end{tabular} & \textbf{80\%}                                   \\ \hline
\end{tabular}
\end{table}

\section{Related Work}\label{related-work}

Federated learning, introduced by McMahan et al.\cite{mcmahan2017communication}, enables learning from decentralized data sources, where clients volunteer to participate in federated learning, i.e., they can join or leave the systems whenever they want.
Simply put, federated learning enables learning from decentralized data sources\cite{kairouz2021advances}.
A variant of federated learning in blockchain settings is swarm learning\cite{warnat2021swarm}, where a smart contract would elect a node to perform model updates at each epoch instead of a central aggregator. This selected node aggregates and broadcasts the model parameters to all other nodes. We drew inspiration from this methodology in the work presented in this paper.
However, swarm learning nodes are essentially large and powerful hospital servers utilized in applications such as leukemia and tuberculosis prediction\cite{warnat2021swarm}. Our work involves learning in a much more decentralized setting that leverages IoT devices and smartphones with much smaller memory and performance. At the same time, input features are all uniform in the original swarm learning implementation\cite{warnat2021swarm}, which we believe is an assumption that may not hold in other decentralized settings. We have embraced the non-IID assumption in our implementation. 

Pfitzner et al.\cite{pfitzner2021federated} conducted a systematic literature review on the concept of and research into federated learning and its applicability for confidential healthcare datasets.
In particular, Lee and Shin\cite{lee2020federated} conducted an experiment using the Modified National Institute of Standards and Technology (MNIST), Medical Information Mart for Intensive Care-III (MIMIC-III), and ECG datasets to evaluate the performance of a federated learning system compared to the state-of-the-art method for in-hospital mortality using imbalanced data. 
Additionally, a small but growing number of works have focused on the application of federated learning in mental health applications. 
FedMood\cite{xu2021fedmood} uses mobile phone and IoT data in a ``multi-view'' federated learning setting to detect the emotions of individuals. However, a central aggregator is still necessary for federated learning to be possible, which is risky, especially for patients in vulnerable populations. 
Chhikara et al.\cite{chhikara2020federated} describe a federated learning framework that uses images to detect human emotion. While they achieved successful learning results, such a model is infeasible for granular changes in emotion/mood. While physiological data can provide hour-by-hour changes in a patient's emotion (such as changes after taking an antidepressant), obtaining patient pictures every hour is infeasible. 
%
Garriga et al.\cite{garriga2022machine}'s work is similar to the previously described work but utilizes EHR to predict mental health crises. While this is a valuable model for predicting specific mental health events, the model cannot assess granular emotional changes. Instead, crises are only extrapolated based on patient visits to the clinic or the hospital (which often occurs when the patient is sick). Through our work, we aim to see both granular and longitudinal changes in emotions, i.e., how mood changes throughout the day (especially with medication interventions) and how moods change over a week. 
A novel variant of federated learning is personalized federated learning\cite{fallah2020personalized}.
The generated model is adapted to better fit a local dataset (for example, data belonging to a single patient). 
Such a model adaptation can lead to a more personalized model for the specific patient. In other words, each client’s model does not need to be the same. While this strategy can prove useful for this application, we have not yet employed it due to problems with categorical overfitting (where the model chooses one category with any given input). Such overfitting will likely occur in this setting as emotion/mood may remain the same for many hours. Subjecting this to a model may lead the model to assume that the patient's emotions always remain the same. A more generalized model can expose a greater variation of training data. 
Architecturally, \emph{MentalHealthAI} provides many learning advantages that other AI strategies in this domain do not provide. Compared to traditional machine learning, \emph{MentalHealthAI} introduces privacy-sensitive strategies to address a previously stigmatized population. We specifically allow learning on decentralized edge nodes, i.e., smartphones and only require transferring model parameters to make \emph{MentalHealthAI} less susceptible to noise and irrelevant features. However, \emph{MentalHealthAI} takes this further by introducing decentralized aggregators, preventing attacks on a centralized aggregator. The unique contribution of \emph{MentalHealthAI} lies in its ability to utilize the available data, regardless of variations in the feature set. 

\section{Conclusion and Future Work}
\label{sec:future-work}

Advances in AI techniques and IoT devices have transformed how chronic illnesses are treated today, such as asthma, hypertension, and diabetes. However, an area of medicine where connected health has remained relatively untapped has been mental health. In most situations, patient history, which is inaccurate and imperfect, has predominantly been used to treat this disorder. At the same time, psychiatry poses many unique challenges to connected health adoption. First, a sensitive patient population may not support releasing personal data, i.e., information potentially harmful if leaked. Secondly, significant variations exist in the number of data streams a patient has, thus potentially limiting learning. Finally, qualitative elements such as mood or emotions can change rapidly throughout the day and the week. For example, simple changes in diet, medications, or even sleep can lead to different emotions. Therefore, monitoring granular emotional changes is key to successfully monitoring mental health. While many have focused on using facial expressions or EHR records to predict crises, these models overlook the small changes that can lead to mental health issues. Therefore, to solve these problems, we utilized a unique combination of various AI and blockchain techniques to enhance data privacy and ownership in a system called \emph{MentalHealthAI}.
It is an innovative combination of smart contracts and decentralized learning to create models useful for psychiatrists but in a way that protects the patient's privacy. IoT devices provide a second-by-second change in the patient's outward physiological signs, allowing for a granular understanding of the patient's health. It allows for successful learning even when data is stored in a patient's smartphone. 
As part of the evaluation, we used a novel mental health dataset and divided the patient population into cohorts based on the data streams available for each patient. 
We demonstrated that we could predict emotions/moods from physiological data in a decentralized and privacy-preserving manner. 

Our methodology for predicting mental health disorders has several benefits. First, it increases accessibility. For example, if 20\% of the patient population has only one IoT device, a traditional machine learning algorithm would be trained on only this limited population. In comparison, \emph{MentalHealthAI} can utilize the entire patient population for model training in a decentralized and privacy-preserving manner, which can provide greater model utility for patients who do not have access to physiological data generators (i.e., IoT Devices). Secondly, it can increase model accuracy, especially in fields that have yet to be studied extensively due to non (or limited) data availability. Therefore, certain data streams may contribute to the model's accuracy, and different combinations of data stream features enable the better establishment of links between features and labels. Finally, this method can better adapt to non-IID settings. Intuitively, patient populations are unique, as most patients are more likely to have between 1-3 IoT devices. Therefore, a model trained on more patients but with fewer features can have greater accuracy than one trained on more features but with a smaller patient cohort. This relationship can change from community to community and region to region. We address this issue by having different population cohorts to provide accurate results while being resilient to changes in the patient population composition. 

There are various limitations to this work. We are yet to evaluate our work with real smartphones in a decentralized setting in real life. Therefore, an initial user study is necessary to determine the effectiveness and impact of the prediction accuracy. Secondly, understanding physiological changes during emotions such as surprise, fear, or agitation would be valuable in addition to detecting moods and emotions in patients as a baseline. Therefore, the current model may need to be retrained on a separate dataset, and hyperparameters tuned appropriately to recognize these emotions. 
Another important challenges are model approximation and optimization, i.e., is there a model that performs well on all clients? And how to find such a model?
By continuing to work on these limitations, we can deploy such infrastructure in the mental health patient population and provide utility to psychiatrists needing an objective metric to assess their patients. 


\setlength{\bibsep}{0pt plus 0.3ex}
\bibliographystyle{ACM-Reference-Format}
\bibliography{references}

\end{document}